\pdfoutput=1

\documentclass[11pt]{article}

\usepackage[]{EACL2023}

\usepackage{times}
\usepackage{latexsym}

\usepackage[utf8]{inputenc}
\usepackage[greek, english]{babel}
\usepackage[LGR, T1]{fontenc}
\usepackage{alphabeta}
\usepackage{booktabs}

\usepackage[utf8]{inputenc}

\usepackage{microtype}

\usepackage{inconsolata}
\usepackage{amsmath}
\usepackage{svg}
\usepackage{graphicx}
\usepackage{xcolor}

\newcommand{\spar}{$S_{\text{parallel}}$ }
\newcommand{\sdif}{$S_\text{different}$ }
\newcommand{\sparn}{$S_{\text{parallel}}$}
\newcommand{\sdifn}{$S_\text{different}$}

\usepackage{todonotes}
\makeatletter
\newcommand*\iftodonotes{\if@todonotes@disabled\expandafter\@secondoftwo\else\expandafter\@firstoftwo\fi}  
\makeatother

\definecolor{myred}{RGB}{163, 11, 11}
\definecolor{myblue}{RGB}{51,141,204}

\title{Multilingual BERT has an accent: \\ Evaluating English influences on fluency in multilingual models}

\author{Isabel Papadimitriou* \and Kezia Lopez* \and Dan Jurafsky \\
    Computer Science Department \\
  Stanford University \\
  \texttt{\{isabelvp,keziakl,jurafsky\}@stanford.edu} 
}
\begin{document}
\maketitle

\begin{abstract}
While multilingual language models can improve NLP performance on low-resource languages by leveraging higher-resource languages, they also reduce average performance on all languages (the `curse of multilinguality').  Here we show another problem with multilingual models: grammatical structures in higher-resource languages bleed into lower-resource languages, a phenomenon we call {\em grammatical structure bias}.  We show this bias via a novel method  for comparing the fluency of multilingual models to the fluency of monolingual Spanish and Greek models: testing their preference for two carefully-chosen variable grammatical structures (optional pronoun-drop in Spanish and optional Subject-Verb ordering in Greek). We find that multilingual BERT is biased toward the English-like setting (explicit pronouns and Subject-Verb-Object ordering) 
as compared to our monolingual control language model.  With our case studies, we hope to bring to light the  fine-grained ways in which multilingual models can be biased,
and encourage more linguistically-aware fluency evaluation.
\end{abstract}

\section{Introduction}
Multilingual language models share a single set of parameters between many languages, opening new pathways for multilingual and low-resource NLP.
However, not all training languages have an equal amount, or a comparable quality of training data in these models. In this paper, we investigate if the hegemonic status of English influences other languages in multilingual language models. We propose a novel method for evaluation, whereby we ask if model predictions for lower-resource languages exhibit structural features of English. This is similar to asking if the model has learned some languages with an ``English accent'', or an English \textit{grammatical structure bias}.

\begin{figure}[t]
    \centering
    \includegraphics[width=\columnwidth]{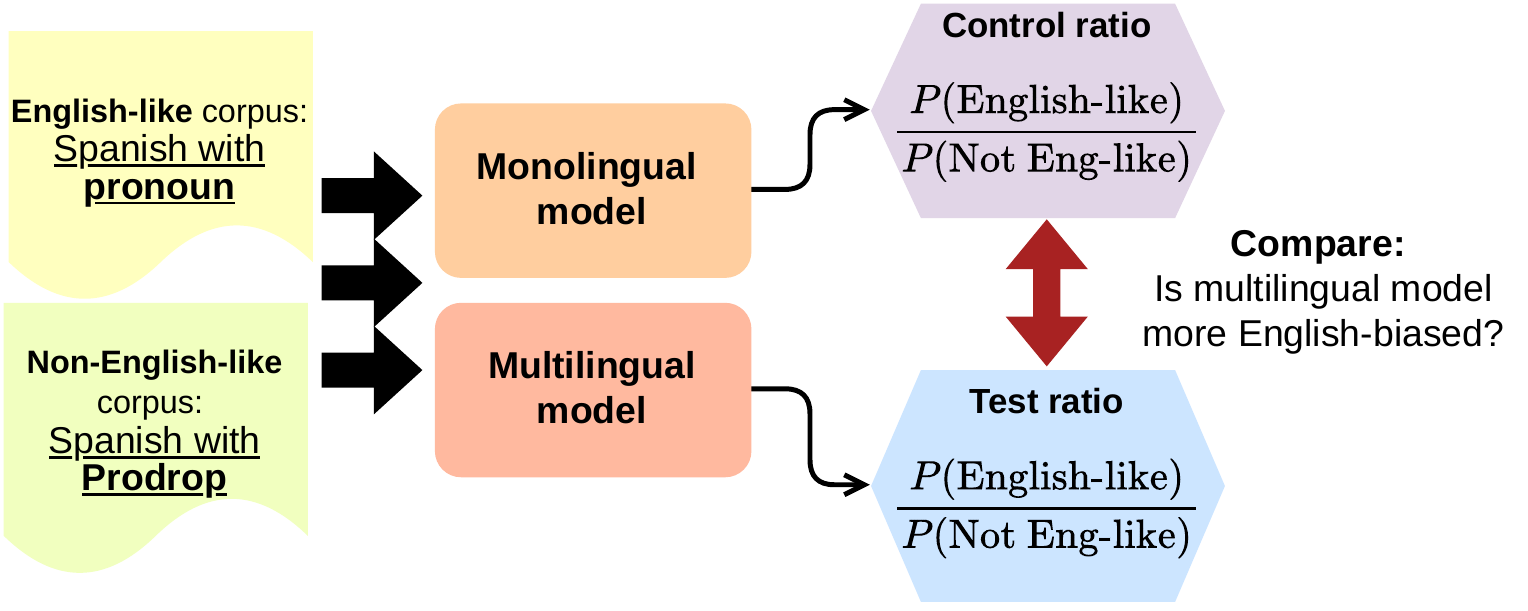}
    \caption{Our method for evaluating English structural bias in multilingual models. We compare monolingual and multilingual model predictions on two sets of natural sentences in the target language: one which is structurally parallel to English, and one which is not.}
    \label{fig:my_label}
\end{figure}

We demonstrate this bias effect in Spanish and Greek, comparing the monolingual models BETO \citep{beto} and GreekBERT \citep{greekbert} to multilingual BERT (mBERT), where English is the most frequent language in the training data. We show that \textit{mBERT 
prefers English-like sentence structure in Spanish and Greek} compared to the monolingual models. Our case studies focus on Spanish pronoun drop (pro-drop) and Greek subject-verb order, two structural grammatical features. We show that multilingual BERT is structurally biased towards explicit pronouns rather than pro-drop in Spanish, and subject-before-verb order in Greek: the structural forms parallel to English.

\begin{table*}[t]
    \centering
    \begin{tabular}{p{\columnwidth} p{0.98\columnwidth}}
    \toprule
    \multicolumn{1}{c}{\large{\textbf{\sparn: English-like structure }}} & \multicolumn{1}{c}{\large{\textbf{\sdifn: Different structure}}} \\
    \toprule
    \textbf{Spanish explicit pronoun} \small{(pron in red, verb in blue)} &
    \textbf{Spanish prodrop} \small{(verb in blue)}\\
    \midrule
         \textcolor{myred}{Yo} \textcolor{myblue}{\underline{volveré}} para averiguarlo&
         Jamás \textcolor{myblue}{\underline{dan}} soluciones y siempre [\dots] \\ 
         \hspace{0.5cm} \small{\textcolor{myred}{I} \textcolor{myblue}{will return} to figure it out} & 
         \hspace{0.5cm}\small{[They] Never \textcolor{myblue}{give} solutions and always [\dots]}
         \\
         El 2004 , \textcolor{myred}{ella} \textcolor{myblue}{\underline{hizo}} doblaje a el Inglés [\dots] & 
         \textcolor{myblue}{\underline{Jugó}} de centrocampista en el Real Zaragoza\\
         \hspace{0.5cm}\small{In 2004, \textcolor{myred}{she} \textcolor{myblue}{did} dubbing to English [\dots]}&
         \hspace{0.5cm}\small{[He/She/You] \textcolor{myblue}{Played} as a midfielder in Real Zaragoza}\\
         \textcolor{myred}{Ella} \textcolor{myblue}{\underline{decide}} pasar sus vacaciones en la granja &
         \textcolor{myblue}{\underline{Habita}} en Perú . \\
         \hspace{0.5cm}\small{\textcolor{myred}{She} \textcolor{myblue}{decides} to spend her vacation in the country} & \hspace{0.5cm}\small{[He/She/You] \textcolor{myblue}{Lives} in Peru}\\
        \toprule 

         \textbf{Greek Subject-Verb} \small{(subject in red, verb in blue)} &
         \textbf{Greek Verb-Subject} \small{(subject in red, verb in blue)}\\
         \midrule
        \textcolor{myred}{\underline{Πηγές}} της Αντιπολίτευσης \textcolor{myblue}{αναφέρουν} ότι [\dots] &
         Το σκορ του αγώνα \textcolor{myblue}{\underline{άνοιξε}} ο \textcolor{myred}{Γουέν} Ρούνι\\
         \hspace{0.5cm}\small{\textcolor{myred}{Sources} of the Opposition \textcolor{myblue}{mention} that [\dots]}& 
         \hspace{0.5cm}\small{The score of the game \textcolor{myblue}{opened} \textcolor{myred}{Wayne} Rooney}\\
        Σε άλλες πλευρές ο \textcolor{myred}{\underline{ποταμός}} \textcolor{myblue}{κυλά} από ψηλούς βράχους &
         Εδώ πρέπει να \textcolor{myblue}{\underline{γίνουν}} μεγαλύτερες \textcolor{myred}{προσπάθειες}. \\
        \hspace{0.5cm}\small{On other sides, the \textcolor{myred}{river} \textcolor{myblue}{flows} from tall boulders}&
         \hspace{0.5cm}\small{Here must \textcolor{myblue}{happen} bigger  \textcolor{myred}{efforts}}\\
        Η \textcolor{myred}{\underline{εκπαίδευση}} και η μόρφωση \textcolor{myblue}{απέκτησαν} επιτέλους προτεραιότητα &
         Απασχόληση στο εξωτερικό \textcolor{myblue}{\underline{ψάχνουν}} οι μισοί \textcolor{myred}{Έλληνες} σε παραγωγική ηλικία\\
         \hspace{0.5cm}\small{\textcolor{myred}{Training} and education have finally \textcolor{myblue}{acquired} priority}&
         \hspace{0.5cm}\small{Employment in foreign countries \textcolor{myblue}{search} half of \textcolor{myred}{Greeks}} \\
         \bottomrule
    \end{tabular}
    \caption{Examples from our dataset for \spar and \sdif in Spanish and Greek, along with roughly word-by-word gloss translations in English. In all cases, we've underlined $w(x)$, the word we use to represent the construction in our calculations. These examples are not randomly selected and have been chosen to be significantly shorter than the average sentence in our datasets in order to be presentable in a table.}
    \label{tab:my_label}
\end{table*}

Though the effect we showcase here is likely not captured by the downstream classification tasks often used to evaluate multilingual models \cite{hu2020xtreme}, it demonstrates the type of fluency that can be lost with multilingual training
--- something that current evaluation methods miss. 
In fact, though we choose two clear-cut syntactic features to investigate, there are many less-measurable features that make language production fluent: subtleties in lexical choice, grammatical choice, and discourse expression, among many others.
With this paper, beyond showing a trend for two specific grammatical features, we wish to highlight fluency discrepancies in multilingual models, and also call for more evaluations focused on fluency.

Our proposed method can be expanded, without the need for manual data collection, to any language with a syntactic treebank and a monolingual model. Since our method focuses on fine-grained linguistic features, some expert knowledge of the target language is necessary for evaluation. Multilingual evaluation so far has been largely translated or automatically curated, and the methods for creating such datasets have allowed for the creation of resources in many languages for which there there were none. Fluency evaluation requires some linguistic expertise to set up, and as such is more restricted in the languages the research community can reach. 
Nevertheless, such evaluation has been missing from the multilingual NLP literature, and our work bridges this gap by proposing fluency testing for multilingual models. 


Our work builds off of a long literature on multilingual evaluation which has until now mostly focused on downstream classification tasks \cite{conneau2018xnli, ebrahimi2022americasnli, clark2020tydiqa, liang2020xglue, hu2020xtreme, raganato2020xlwic, li2021mtop}. With the help of these evaluation methods, research has pointed out the problems for both high- and low-resource languages that come with adding many languages to a single model \citep[][inter alia]{wang2020negative, turc2021primacy, lauscher2020zero}. 
Methods for creating more equitable models  have been proposed, through identifying or reserving language-specific parameters for each language \cite{ansell2022composable, pfeiffer2022lifting}, through training models without  tyoplogically distant languages that dominate the training data \citep{afriberta, virtanen2019finnish, ogunremi2023mini}, as well as through adding model capacity \citep{conneau2019unsupervised, xue2021mt5, lepikhin2021gshard, liang2023xlmv}. We hope that our work can add to these analyses and methodologies by pointing out issues beyond downstream classification performance that can arise with multilingual training, and aid towards building and evaluating more equitable multilingual models.




\section{Method}

Our method relies on finding a variable construction in the target language which can take two structural surface forms: one which is parallel to English (\sparn) and one which is not (\sdifn). Surface forms parallel to English are those which mirror English structure. For example, English has strict Subject-Verb-Object word order, and so a \textit{parallel} structure in another language is one where the verb and its arguments appear in Subject-Verb-Object order, while a \textit{different} structure is one where the verb appears before the subject (see Table \ref{tab:my_label} for examples). 

Once we have identified such a construction in our target language, we can ask: are multilingual models biased towards \sparn? For a native speaker of the target language, structural, semantic, and discourse features determine whether they will use \spar or \sdif in a given context --- with the alternative option usually being less fluent. We assume that a BERT-sized monolingual model in the target language will have a sufficiently accurate representation of this fluent variation between \spar and \sdif without being influenced by other languages. Therefore, to understand if multilingual models have an English structural bias, we now just have to answer: do multilingual models prefer \spar over \sdif \textit{more} than the fluent distribution defined by a monolingual model?


\subsection{Collecting model judgements}
By design, both \spar and \sdif are constructions that occur naturally in the target language. Therefore, we should be able to use the syntactic treebank annotations to pick out sentences that exhibit the structures \spar or \sdifn. We can put these extracted sentences into two corpora, $C_\text{parallel}$ and $C_\text{different}$. Note that the sentences in $C_\text{parallel}$ and $C_\text{different}$ are unrelated and not paired, and that the two corpora can have different sizes. 
Crucially, we have to use natural sentences for both of our corpora: we cannot artificially alter sentences from \spar to \sdifn, or use templates to create sentences. This is because our evaluation is about the subtleties of fluency, while altered or templated stimuli are not naturally produced and are therefore often awkward, confounding any effect we might want to measure.  

Each model gives us a ratio $r_\mathtt{model}$: the average probability of a sentence in $C_\text{parallel}$ divided by the average probability of a sentence in $C_\text{different}$ according to the model. That is: 
\begin{align} \label{eq:rmodel}
r_\mathtt{model} = \dfrac{\sum_{x \in C_p}P_\mathtt{model}(x) \mathbin{/} |C_p|}{\sum_{x \in C_d}P_\mathtt{model}(x) \mathbin{/} |C_d|}    
\end{align}

We want to compare judgements on these corpora from two models: a monolingual model $\mathtt{mono}$ and a multilingual model $\mathtt{multi}$. Our experimental question then boils down to asking if $r_\mathtt{multi}$ is significantly larger than $r_\mathtt{mono}$. 

\subsection{From model outputs to construction probability}
How can we calculate $P_\mathtt{model}(x)$ for a given sentence $x$, focusing on the probability of a specific construction in $x$? Looking at model judgements over long natural sentences introduces a lot of noise that is unrelated to the structural construction in question, reducing the statistical power of our experiment. Furthermore, since we are looking at encoder-only bidirectional models, there is no canonical or controlled way of extracting the probability of a whole sentence. To get a better model judgement for each sentence, we can extract the probability of \textit{one word} in each sentence that best represents the construction. For example, if we are looking at pronoun drop, it makes sense to use main verb of the sentence as the target word, as this is the syntactic head of the pronoun that is present or dropped. Using a carefully chosen word as a proxy for the probability of a construction is a methodological choice also made in reading time psycholinguistics experiments \citep{levy2011integrating, levy2013expectation}. 

Going back to our problem of calculating $P_\mathtt{model}(x)$, we define $w$ to be a function that returns the structurally-relevant word from each sentence. Using this, we approximate $P_\mathtt{model}(x)$ in Eq. (\ref{eq:rmodel}) with $P_\mathtt{model}(w(x) | x)$. 
The probability $P(w(x) | x)$ is simple to calculate for BERT-style masked language models: it is simply the logit of the word $w(x)$ when we encode the sentence $x$ using $\mathtt{model}$.

\subsection{Extending to more languages}
Extending our fluency evaluation to a new language requires three language-specific steps: (1) decide on an appropriate construction with two structural forms \spar and \sdifn, (2) decide on an appropriate $w(x)$: which word in each structural form can represent the form, and (3) use treebank annotations to pull out sentences which exhibit \spar or \sdifn, and identify the relevant word. Below, we detail these steps for our two case studies.

\begin{figure}[t]
    \centering
    \includegraphics[trim = {5.5cm 0cm 5.5cm 0cm}, clip, width=\columnwidth]{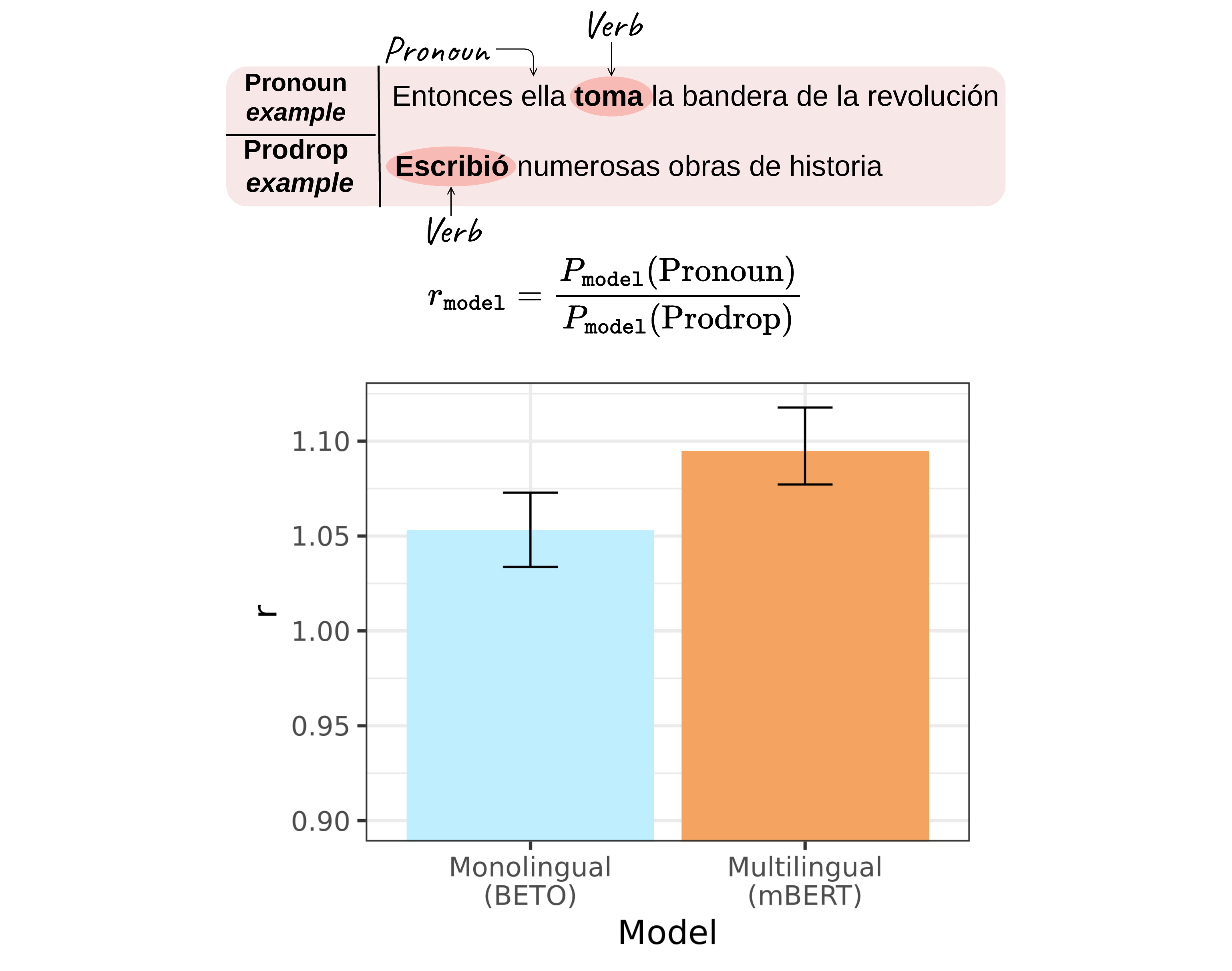}
    \caption{Results from our experiment on the Spanish GSD treebank, along with two examples from the treebank to illustrate \spar (with pronoun) and \sdif (pro-drop). We compare model logits for the main verb of the sentence, which is bold and highlighted in the examples. Error bars represent 95\% bootstrap confidence intervals. We find that $r_\mathtt{mono}$ is significantly smaller than $r_\mathtt{multi}$ (bootstrap sampling, $p<0.05$). }
    \label{fig:spanishresults}
\end{figure}

\subsection{Case Study: Spanish Pro-drop}

In Spanish, the subject pronoun is often dropped: person and number are mostly reflected in verb conjugation, so the pronoun is realized or dropped depending on semantic and discourse factors. English, on the other hand, does not allow null subjects except in rare cases, and expletive syntactic subjects like ``there'' are even added when there is no clear subject. For our Spanish experiment, we define \spar to be sentences which have the subject pronoun of the main verb, as is necessary in English, and \sdif to be pro-drop sentences which have a main verb with no realized subject. We define $w$ to be the main verb of the sentence, which is always present in our extracted examples. 

To extract our corpora $C_\text{parallel}$ and $C_\text{different}$, we use the Spanish GSD treebank from the Universal Dependencies dataset \citep{demarneffe2021universal}. We ignore all sentences not verb-rooted (i.e. noun phrases), those rooted with ``haber'' (which in its copula-like existential form cannot take an explicit subject, ``There is'' in English), and those using the impersonal-``se'' passive construction (e.g. ``se nos fue permitido'', ``it was permitted of us''). We then take all sentences with a pronoun subject (i.e. a pronoun dependent of the root verb) and add them to $C_\text{parallel}$ and all sentences where there is no \texttt{nsubj} relation to root verb and add them to $C_\text{different}$. We always pick the main root verb of the sentence as our $w$. We collect 283 sentences in $C_\text{parallel}$ and 2,656 sentences in $C_\text{different}$.

\subsection{Case Study: Greek Subject-Verb order}

English is a fixed word order language: with few exceptions, the order of a verb and its arguments is Subject-Verb-Object. Greek, on the other hand, has mostly free word order \citep{mackridge1985greek}, meaning that the verb and arguments can appear in any order that is most appropriate given discourse context.
For our experiment, we define \spar to be cases in Greek when the subject precedes the verb, as is the rule in English. \sdif is then the cases when the verb precedes the subject, which almost never happens in English. 

We define $w$ to be the first element of the subject and verb: the subject when the subject comes first or the verb when the verb comes first. This first element is closer to the surrounding context, and so gives us a word-order-sensitive measurement of how the subject-verb construction is processed as a whole within the context.
Though this choice means that our $w$ is a noun in \spar and a verb in \sdifn, this does not constitute a confounder between models: we are comparing the same noun-verb probability ratio between different models. 



To extract our corpora $C_\text{parallel}$ and $C_\text{different}$, we use the Greek Dependency Treebank, the Universal Dependencies treebank for Greek  \cite{prokopidis2017gdt}. We take all sentences where the main verb has a lexical subject, and we add to $C_\text{parallel}$ if the subject appears before the verb and to $C_\text{different}$ if it appears after.  We collect 1,446 sentences in $C_\text{parallel}$ and 425 sentences in $C_\text{different}$.

\begin{figure}[t]
    \includegraphics[trim = {7cm 0cm 7cm 0cm}, clip, width=\columnwidth]{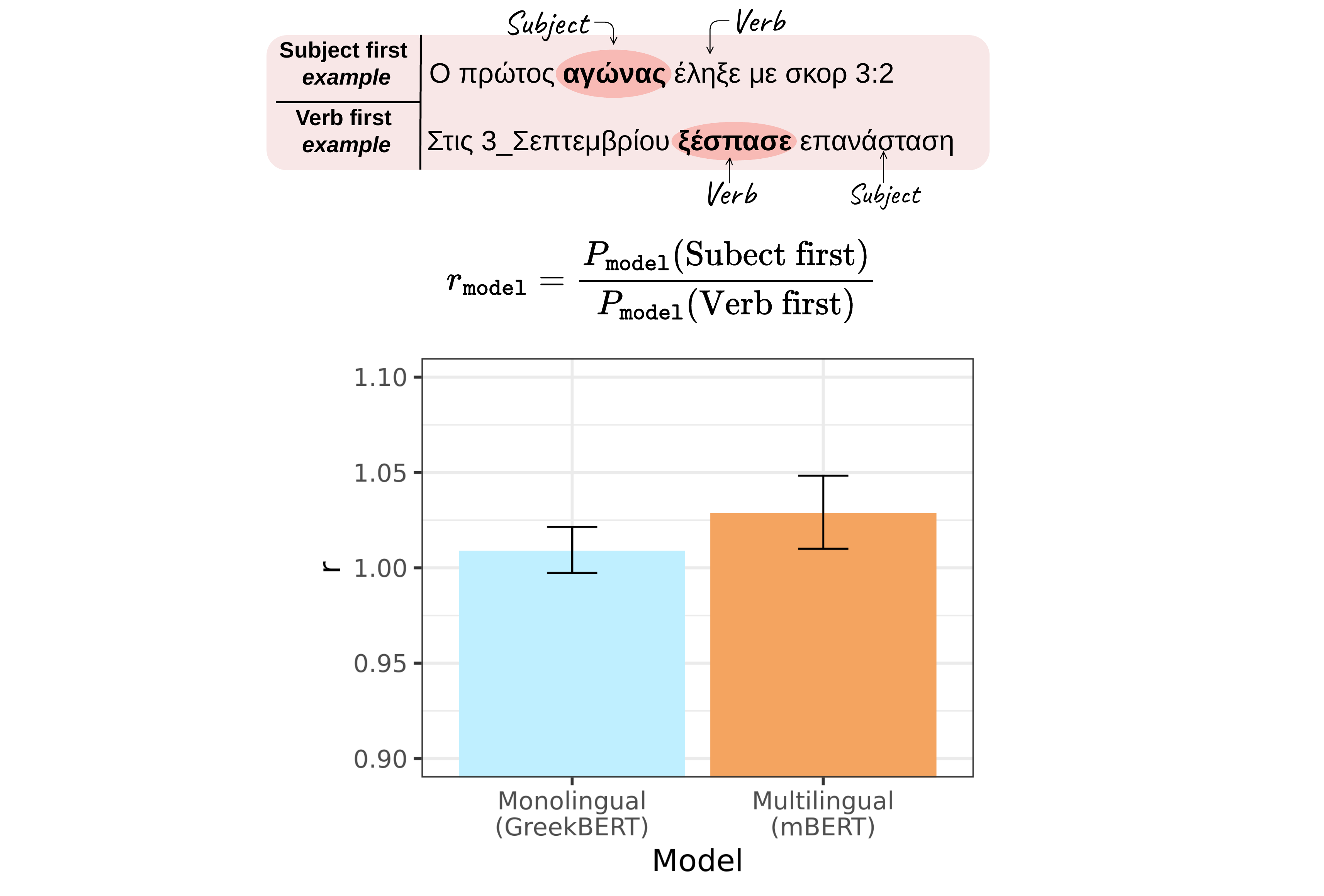}
    \caption{Results from our experiment on the Greek Dependency Treebank, along with two examples from the treebank to illustrate \spar (Subject-Verb) and \sdif (Verb-Subject). We measure and compare model logits for the bold words: the subject in subject-verb sentences and the verb in verb-subject sentences. Error bars represent 95\% bootstrap confidence intervals. $r_\mathtt{mono}$ is significantly smaller than $r_\mathtt{multi}$ (bootstrap sampling, $p< 0.05$).}
    \label{fig:greekresults}
\end{figure} 

\section{Results}

Results are shown in Figures \ref{fig:spanishresults} and \ref{fig:greekresults}, showing for both of our case studies that multilingual BERT has a greater propensity for preferring English-like sentences which exhibit \sparn. Multilingual BERT significantly prefers pronoun sentences over pro-drop compared with monolingual BETO (bootstrap sampling, $p< 0.05$), and significantly prefers subject-verb sentences over verb-subject sentences over GreekBERT (bootstrap sampling, $p < 0.05$).


\section{Discussion}

In this paper, we proposed fluency evaluation as a further way of understanding the curse of multilinguality: what can be lost when we train many languages together. The discrepancies that we point out in these experiments are not going to seriously affect multilingual LM performance, especially in the more coarse-grained classification tasks that are most commonly used for evaluation.
But, as we demonstrate here, not all levels of language learning can be evaluated from such datasets.

Our experiments do not pinpoint the reasons behind the effects that we measure: there are different possible explanations for the English-like trends that we showcase. On the one hand, the effects we measure might stem from training with a language that's more dominant in the training data, like English is for many multilingual models. Such training could lead to an English-biased representation space which the representations of other languages conform to. On the other hand, the effects we show might be down to the data: the non-English datasets used to train a multilingual model may be more limited in domain, may contain a high proportion of data that's actually been translated from English \citep[Multilingual Wikipedia is often translated,][]{adar2009arbitrage}, or might be more polluted with irrelevant or non-linguistic elements. Domain limitations and translationese stemming from the data are separate, but related issues to fluency: fluency can be grammatical, but also involves proficiency in a range of registers or possibilities. 
It is also possible that the effects we show are due to a combination of both multilingual representation learning artifacts, and training data quality. 
Further controlled fluency experimentation on the limits and abilities of multilingual models is needed to disentangle these effects. 
We hope the case studies in this paper can inspire more fine-grained evaluation of multilingual models, so that we understand the ``accent''-like effects of hegemonic languages more fully.

\section{Limitations}

This study is meant to highlight the kinds of modeling flaws that have so far gone undetected and that can arise for lower-resource languages in multilingual models. However, our study does not focus on languages that are truly low-resource. In fact, as designed it could not do so: our methodology relies on having an available monolingual model, which of course requires a large amount of training data. This is because our method requires a control: we can only judge multilingual models against what we can believe to be a non-biased language model in the language. There are ways to test for fluency in low-resource languages that would not require a monolingual model as a control, but would require dataset collection in the target language for features that reflect fluency and linguistic acceptability (similar to what \citet{warstadt2019cola} achieve with the CoLA dataset for English). We hope our study can create motivation for such work in linguistically-aware, fine-grained multilingual evaluation for languages of all resource levels.

Our experiments focus on BERT-style models, since this is mostly the size of model available for monolingual, non-English models (in our case BETO and GreekBERT). However, it is not necessary from these experiments that our findings extrapolate to larger models that are commonplace at the time of writing. 

Lastly, both pro-drop and subject-verb order are largely discourse-dependent constructions. For example, pro-drop is more likely when the subject of the sentence is very clear from the discourse, while subject-verb order in Greek is changed to achieve different discourse focus, similar to how intonation changes the focus of a sentence in English (e.g., stressing the verb in ``Mary \textit{helped} John'' puts the focus on the verb, which in Greek can be done by putting the verb first). Despite this, all of our experiments are done on isolated sentences from the UD treebanks and do not contain discourse content. Though this means that the models do not have the full relevant context for each input, we do not expect that having more context should favor one model more than another for our evaluation. Since this work compares models on the same inputs, we did not consider this a significant confounder. 

\section{Acknowledgements}

We thank Anjalie Field and Mirac Suzgun for comments on drafts. 
This research was funded in part by NSF award number IIS-2128145. 

\bibliography{anthology,custom}

\begin{thebibliography}{28}
\expandafter\ifx\csname natexlab\endcsname\relax\def\natexlab#1{#1}\fi

\bibitem[{Adar et~al.(2009)Adar, Skinner, and Weld}]{adar2009arbitrage}
Eytan Adar, Michael Skinner, and Daniel~S. Weld. 2009.
\newblock \href {https://doi.org/10.1145/1498759.1498813} {Information
  arbitrage across multi-lingual wikipedia}.
\newblock In \emph{Proceedings of the Second ACM International Conference on
  Web Search and Data Mining}, WSDM '09, page 94–103, New York, NY, USA.
  Association for Computing Machinery.

\bibitem[{Ansell et~al.(2022)Ansell, Ponti, Korhonen, and
  Vuli{\'c}}]{ansell2022composable}
Alan Ansell, Edoardo Ponti, Anna Korhonen, and Ivan Vuli{\'c}. 2022.
\newblock Composable sparse fine-tuning for cross-lingual transfer.
\newblock In \emph{Proceedings of the 60th Annual Meeting of the Association
  for Computational Linguistics (Volume 1: Long Papers)}, pages 1778--1796.

\bibitem[{Cañete et~al.(2020)Cañete, Chaperon, Fuentes, Ho, Kang, and
  Pérez}]{beto}
José Cañete, Gabriel Chaperon, Rodrigo Fuentes, Jou-Hui Ho, Hojin Kang, and
  Jorge Pérez. 2020.
\newblock Spanish {P}re-trained {BERT} {M}odel and {E}valuation {D}ata.
\newblock In \emph{PML4DC at ICLR 2020}.

\bibitem[{Clark et~al.(2020)Clark, Choi, Collins, Garrette, Kwiatkowski,
  Nikolaev, and Palomaki}]{clark2020tydiqa}
Jonathan~H Clark, Eunsol Choi, Michael Collins, Dan Garrette, Tom Kwiatkowski,
  Vitaly Nikolaev, and Jennimaria Palomaki. 2020.
\newblock {TyDi QA: A} benchmark for information-seeking question answering in
  typologically diverse languages.
\newblock \emph{Transactions of the Association for Computational Linguistics},
  8:454--470.

\bibitem[{Conneau et~al.(2020)Conneau, Khandelwal, Goyal, Chaudhary, Wenzek,
  Guzm{\'a}n, Grave, Ott, Zettlemoyer, and Stoyanov}]{conneau2019unsupervised}
Alexis Conneau, Kartikay Khandelwal, Naman Goyal, Vishrav Chaudhary, Guillaume
  Wenzek, Francisco Guzm{\'a}n, {\'E}douard Grave, Myle Ott, Luke Zettlemoyer,
  and Veselin Stoyanov. 2020.
\newblock Unsupervised cross-lingual representation learning at scale.
\newblock In \emph{Proceedings of the 58th Annual Meeting of the Association
  for Computational Linguistics}, pages 8440--8451.

\bibitem[{Conneau et~al.(2018)Conneau, Rinott, Lample, Williams, Bowman,
  Schwenk, and Stoyanov}]{conneau2018xnli}
Alexis Conneau, Ruty Rinott, Guillaume Lample, Adina Williams, Samuel Bowman,
  Holger Schwenk, and Veselin Stoyanov. 2018.
\newblock {XNLI: Evaluating Cross-lingual Sentence Representations}.
\newblock In \emph{Proceedings of the 2018 Conference on Empirical Methods in
  Natural Language Processing}, pages 2475--2485.

\bibitem[{De~Marneffe et~al.(2021)De~Marneffe, Manning, Nivre, and
  Zeman}]{demarneffe2021universal}
Marie-Catherine De~Marneffe, Christopher~D Manning, Joakim Nivre, and Daniel
  Zeman. 2021.
\newblock Universal {D}ependencies.
\newblock \emph{Computational linguistics}, 47(2):255--308.

\bibitem[{Ebrahimi et~al.(2022)Ebrahimi, Mager, Oncevay, Chaudhary, Chiruzzo,
  Fan, Ortega, Ramos, Gonzales, Meza-Ruiz et~al.}]{ebrahimi2022americasnli}
Abteen Ebrahimi, Manuel Mager, Arturo Oncevay, Vishrav Chaudhary, Luis
  Chiruzzo, Angela Fan, John Ortega, Ricardo Ramos, Annette~Rios Gonzales, Ivan
  Meza-Ruiz, et~al. 2022.
\newblock {AmericasNLI: Evaluating Zero-shot Natural Language Understanding of
  Pretrained Multilingual Models in Truly Low-resource Languages}.
\newblock In \emph{Proceedings of the 60th Annual Meeting of the Association
  for Computational Linguistics (Volume 1: Long Papers)}, pages 6279--6299.

\bibitem[{Hu et~al.(2020)Hu, Ruder, Siddhant, Neubig, Firat, and
  Johnson}]{hu2020xtreme}
Junjie Hu, Sebastian Ruder, Aditya Siddhant, Graham Neubig, Orhan Firat, and
  Melvin Johnson. 2020.
\newblock {XTREME}: A massively multilingual multi-task benchmark for
  evaluating cross-lingual generalisation.
\newblock In \emph{International Conference on Machine Learning}, pages
  4411--4421. PMLR.

\bibitem[{Koutsikakis et~al.(2020)Koutsikakis, Chalkidis, Malakasiotis, and
  Androutsopoulos}]{greekbert}
John Koutsikakis, Ilias Chalkidis, Prodromos Malakasiotis, and Ion
  Androutsopoulos. 2020.
\newblock {Greek-BERT: The Greeks visiting sesame street}.
\newblock In \emph{11th Hellenic Conference on Artificial Intelligence}, pages
  110--117.

\bibitem[{Lauscher et~al.(2020)Lauscher, Ravishankar, Vuli{\'c}, and
  Glava{\v{s}}}]{lauscher2020zero}
Anne Lauscher, Vinit Ravishankar, Ivan Vuli{\'c}, and Goran Glava{\v{s}}. 2020.
\newblock {From Zero to Hero}: On the limitations of zero-shot language
  transfer with multilingual transformers.
\newblock In \emph{Proceedings of the 2020 Conference on Empirical Methods in
  Natural Language Processing (EMNLP)}, pages 4483--4499.

\bibitem[{Lepikhin et~al.(2021)Lepikhin, Lee, Xu, Chen, Firat, Huang, Krikun,
  Shazeer, and Chen}]{lepikhin2021gshard}
Dmitry Lepikhin, HyoukJoong Lee, Yuanzhong Xu, Dehao Chen, Orhan Firat, Yanping
  Huang, Maxim Krikun, Noam Shazeer, and Zhifeng Chen. 2021.
\newblock \href {https://openreview.net/forum?id=qrwe7XHTmYb} {{\{}GS{\}}hard:
  Scaling giant models with conditional computation and automatic sharding}.
\newblock In \emph{International Conference on Learning Representations}.

\bibitem[{Levy(2011)}]{levy2011integrating}
Roger Levy. 2011.
\newblock Integrating surprisal and uncertain-input models in online sentence
  comprehension: {F}ormal techniques and empirical results.
\newblock In \emph{Proceedings of the 49th annual meeting of the Association
  for Computational Linguistics: Human Language Technologies}, pages
  1055--1065.

\bibitem[{Levy and Keller(2013)}]{levy2013expectation}
Roger~P. Levy and Frank Keller. 2013.
\newblock \href {https://doi.org/https://doi.org/10.1016/j.jml.2012.02.005}
  {Expectation and locality effects in {G}erman verb-final structures}.
\newblock \emph{Journal of Memory and Language}, 68(2):199--222.

\bibitem[{Li et~al.(2021)Li, Arora, Chen, Gupta, Gupta, and
  Mehdad}]{li2021mtop}
Haoran Li, Abhinav Arora, Shuohui Chen, Anchit Gupta, Sonal Gupta, and Yashar
  Mehdad. 2021.
\newblock Mtop: A comprehensive multilingual task-oriented semantic parsing
  benchmark.
\newblock In \emph{Proceedings of the 16th Conference of the European Chapter
  of the Association for Computational Linguistics: Main Volume}, pages
  2950--2962.

\bibitem[{Liang et~al.(2023)Liang, Gonen, Mao, Hou, Goyal, Ghazvininejad,
  Zettlemoyer, and Khabsa}]{liang2023xlmv}
Davis Liang, Hila Gonen, Yuning Mao, Rui Hou, Naman Goyal, Marjan
  Ghazvininejad, Luke Zettlemoyer, and Madian Khabsa. 2023.
\newblock Xlm-v: Overcoming the vocabulary bottleneck in multilingual masked
  language models.
\newblock \emph{arXiv preprint arXiv:2301.10472}.

\bibitem[{Liang et~al.(2020)Liang, Duan, Gong, Wu, Guo, Qi, Gong, Shou, Jiang,
  Cao, Fan, Zhang, Agrawal, Cui, Wei, Bharti, Qiao, Chen, Wu, Liu, Yang,
  Campos, Majumder, and Zhou}]{liang2020xglue}
Yaobo Liang, Nan Duan, Yeyun Gong, Ning Wu, Fenfei Guo, Weizhen Qi, Ming Gong,
  Linjun Shou, Daxin Jiang, Guihong Cao, Xiaodong Fan, Ruofei Zhang, Rahul
  Agrawal, Edward Cui, Sining Wei, Taroon Bharti, Ying Qiao, Jiun-Hung Chen,
  Winnie Wu, Shuguang Liu, Fan Yang, Daniel Campos, Rangan Majumder, and Ming
  Zhou. 2020.
\newblock \href {https://doi.org/10.18653/v1/2020.emnlp-main.484} {{XGLUE}: A
  new benchmark datasetfor cross-lingual pre-training, understanding and
  generation}.
\newblock In \emph{Proceedings of the 2020 Conference on Empirical Methods in
  Natural Language Processing (EMNLP)}, pages 6008--6018, Online. Association
  for Computational Linguistics.

\bibitem[{Mackridge(1985)}]{mackridge1985greek}
P.~Mackridge. 1985.
\newblock \href {https://books.google.com/books?id=RWViAAAAMAAJ} {\emph{The
  {M}odern {G}reek Language: A Descriptive Analysis of Standard Modern Greek}}.
\newblock Oxford University Press.

\bibitem[{Ogueji et~al.(2021)Ogueji, Zhu, and Lin}]{afriberta}
Kelechi Ogueji, Yuxin Zhu, and Jimmy Lin. 2021.
\newblock \href {https://doi.org/10.18653/v1/2021.mrl-1.11} {Small {D}ata? {N}o
  {P}roblem! {E}xploring the viability of pretrained multilingual language
  models for low-resourced languages}.
\newblock In \emph{Proceedings of the 1st Workshop on Multilingual
  Representation Learning}, pages 116--126, Punta Cana, Dominican Republic.
  Association for Computational Linguistics.

\bibitem[{\`{O}g\'{u}nr\d{\`{e}}m\'{i} and Manning(2023)}]{ogunremi2023mini}
Tol\'{u}l\d{o}p\d{\'{e}} \`{O}g\'{u}nr\d{\`{e}}m\'{i} and Christopher~D.
  Manning. 2023.
\newblock {M}ini but {M}ighty: {E}fficient multilingual pretraining with
  linguistically-informed data selection.
\newblock In \emph{Findings of EACL 2023}.

\bibitem[{Pfeiffer et~al.(2022)Pfeiffer, Goyal, Lin, Li, Cross, Riedel, and
  Artetxe}]{pfeiffer2022lifting}
Jonas Pfeiffer, Naman Goyal, Xi~Lin, Xian Li, James Cross, Sebastian Riedel,
  and Mikel Artetxe. 2022.
\newblock \href {https://doi.org/10.18653/v1/2022.naacl-main.255} {Lifting the
  curse of multilinguality by pre-training modular transformers}.
\newblock In \emph{Proceedings of the 2022 Conference of the North American
  Chapter of the Association for Computational Linguistics: Human Language
  Technologies}, pages 3479--3495, Seattle, United States. Association for
  Computational Linguistics.

\bibitem[{Prokopidis and Papageorgiou(2017)}]{prokopidis2017gdt}
Prokopis Prokopidis and Haris Papageorgiou. 2017.
\newblock \href {http://www.aclweb.org/anthology/W17-0413.pdf} {Universal
  {D}ependencies for {G}reek}.
\newblock In \emph{Proceedings of the NoDaLiDa 2017 Workshop on Universal
  Dependencies (UDW 2017)}, pages 102--106, Gothenburg, Sweden. Association for
  Computational Linguistics.

\bibitem[{Raganato et~al.(2020)Raganato, Pasini, Camacho-Collados, and
  Pilehvar}]{raganato2020xlwic}
Alessandro Raganato, Tommaso Pasini, Jose Camacho-Collados, and Mohammad~Taher
  Pilehvar. 2020.
\newblock {XL-WiC}: A multilingual benchmark for evaluating semantic
  contextualization.
\newblock In \emph{Proceedings of the 2020 Conference on Empirical Methods in
  Natural Language Processing (EMNLP)}. The Association for Computational
  Linguistics.

\bibitem[{Turc et~al.(2021)Turc, Lee, Eisenstein, Chang, and
  Toutanova}]{turc2021primacy}
Iulia Turc, Kenton Lee, Jacob Eisenstein, Ming-Wei Chang, and Kristina
  Toutanova. 2021.
\newblock \href {https://arxiv.org/abs/2106.16171} {Revisiting the {P}rimacy of
  {E}nglish in {Z}ero-shot {C}ross-lingual {T}ransfer}.
\newblock \emph{CoRR}, abs/2106.16171.

\bibitem[{Virtanen et~al.(2019)Virtanen, Kanerva, Ilo, Luoma, Luotolahti,
  Salakoski, Ginter, and Pyysalo}]{virtanen2019finnish}
Antti Virtanen, Jenna Kanerva, Rami Ilo, Jouni Luoma, Juhani Luotolahti, Tapio
  Salakoski, Filip Ginter, and Sampo Pyysalo. 2019.
\newblock Multilingual is not enough: {BERT} for {F}innish.
\newblock \emph{arXiv preprint arXiv:1912.07076}.

\bibitem[{Wang et~al.(2020)Wang, Lipton, and Tsvetkov}]{wang2020negative}
Zirui Wang, Zachary~C Lipton, and Yulia Tsvetkov. 2020.
\newblock On negative interference in multilingual models: Findings and a
  meta-learning treatment.
\newblock In \emph{Proceedings of the 2020 Conference on Empirical Methods in
  Natural Language Processing (EMNLP)}, pages 4438--4450.

\bibitem[{Warstadt et~al.(2019)Warstadt, Singh, and Bowman}]{warstadt2019cola}
Alex Warstadt, Amanpreet Singh, and Samuel~R Bowman. 2019.
\newblock Neural network acceptability judgments.
\newblock \emph{Transactions of the Association for Computational Linguistics},
  7:625--641.

\bibitem[{Xue et~al.(2021)Xue, Constant, Roberts, Kale, Al-Rfou, Siddhant,
  Barua, and Raffel}]{xue2021mt5}
Linting Xue, Noah Constant, Adam Roberts, Mihir Kale, Rami Al-Rfou, Aditya
  Siddhant, Aditya Barua, and Colin Raffel. 2021.
\newblock {mT5}: A massively multilingual pre-trained text-to-text transformer.
\newblock In \emph{NAACL-HLT}.

\end{thebibliography}
\bibliographystyle{acl_natbib}

\end{document}